\documentclass{article}

\usepackage{arxiv}

\usepackage[utf8]{inputenc} 
\usepackage[T1]{fontenc}    
\usepackage[pagebackref,breaklinks,colorlinks]{hyperref}
\usepackage{url}            
\usepackage{booktabs}       
\usepackage{amsfonts}       
\usepackage{nicefrac}       
\usepackage{microtype}      
\usepackage{lipsum}

\usepackage{amssymb}
\usepackage{amsmath}
\usepackage[normalem]{ulem}
\usepackage{multirow}
\usepackage{array}
\usepackage{graphicx}
\usepackage{bm}
\usepackage{caption,subcaption}        
\usepackage{pifont}

\title{pNNCLR: Stochastic Pseudo Neighborhoods for Contrastive Learning based Unsupervised Representation Learning Problems}

\author{
    Momojit Biswas \\
    Jadavpur University \\
    Kolkata, India \\
    \texttt{mb16biswas@gmail.com} \\
    \And
    Himanshu Buckchash \\
    UiT The Arctic University of Norway \\
    Tromsø, Norway \\
    \texttt{himanshu.buckchash@uit.no} \\
    \And
    Dilip K. Prasad \\
    UiT The Arctic University of Norway \\
    Tromsø, Norway \\
    \texttt{dilip.prasad@uit.no} \\
}

\begin{document}
\maketitle

\begin{abstract}
Nearest neighbor (NN) sampling provides more semantic variations than pre-defined transformations for self-supervised learning (SSL) based image recognition problems. However, its performance is restricted by the quality of the support set, which holds positive samples for the contrastive loss. In this work, we show that the quality of the support set plays a crucial role in any nearest neighbor based method for SSL. We then provide a refined baseline (pNNCLR) to the nearest neighbor based SSL approach (NNCLR). To this end, we introduce pseudo nearest neighbors (pNN) to control the quality of the support set, wherein, rather than sampling the nearest neighbors, we sample in the vicinity of hard nearest neighbors by varying the magnitude of the resultant vector and employing a stochastic sampling strategy to improve the performance. Additionally, to stabilize the effects of uncertainty in NN-based learning, we employ a smooth-weight-update approach for training the proposed network. Evaluation of the proposed method on multiple public image recognition and medical image recognition datasets shows that it performs up to 8 percent better than the baseline nearest neighbor method, and is comparable to other previously proposed SSL methods.
\end{abstract}


\section{Introduction}
\label{sec:introduction}
Deep learning is rapidly revolutionizing almost every sector of our society. Off-the-shelf models are being used for feature/representation extraction, and standard models are being fine-tuned for their application to specific problems \cite{adaloglou2023exploring}. To train such models, efficient representation learning methods are required \cite{dclr}. SSL or representation learning provides the backbone networks for many computer vision related tasks such as object detection, segmentation, image or video recognition, etc. \cite{zheng2023application}. Recent developments like NNCLR \cite{nnclr}, SimSiam \cite{simsiam}, Decouplted contrastive learning \cite{dclr}, CLIP \cite{clipradford2021learning}, CAEs \cite{chen2022context}, are good examples of powerful feature extractors, and all employ some standard backbone network like ResNet \cite{he2016deep} or EfficientNet \cite{koonce2021efficientnet}. Labeling the data is a costly operation, on the other hand, the main advantage of SSL models is their ability to learn better generic representations from the unlabelled data \cite{simclr}. Foundational works like SimCLR \cite{simclr} and SimSiam \cite{simsiam} have established that SSL models with slight finetuning (as low as 1\% of the labeled data) can outperform their counterpart supervised models. Another advantage of SSL models is that they provide task-agnostic models which can easily be adapted using transfer learning to multiple kinds of downstream tasks (the tasks which are specialized cases of a larger generic task also known as the pretext task) \cite{gidaris2018unsupervised}.

Earlier models like the non-contrastive models (RotNet \cite{gidaris2018unsupervised}, Jigsaw \cite{noroozi2016unsupervised}) used intelligently designed pretext tasks for providing the self-supervisory signal to the learning algorithm. These signals were based on some independent tasks like verifying the correct rotation \cite{gidaris2018unsupervised}, the correct sequence of frames \cite{misra2016shuffle}, or the correct placement of tiles \cite{noroozi2016unsupervised}. Recently, another branch of SSL methods, called contrastive learning (CL), has shown promising progress \cite{simclr}. With better loss functions and image augmentations, these models have now exceeded the non-contrastive models. These contrastive learning based methods work by pushing closer the similar-looking (positive) samples and pushing apart non-similar class (negative) samples, without actually knowing the classes of these samples. However, recently, it was shown that the positives generated using augmentations are not very semantically diverse \cite{nnclr}. To overcome this, it was suggested to use the nearest neighbors of the anchors (the samples whose positive is to be found), since this leads to better representations by learning from non-trivial positive samples \cite{nnclr}. However, in our experiments and analysis with NNCLR \cite{nnclr}, we found that the quality of the support set plays a crucial role in learning better representations. At the beginning of training, the probability of finding a good nearest neighbor is low, and this can affect the overall learning in the SSL model. Based on these observations, this work presents stochastic pseudo nearest neighbors and the learning framework --- pNNCLR.

The main objective of contrastive SSL methods is to employ a strategy or function, $f$, to arrange the latent space in such a way that similar class samples appear closer (attraction property) than distinct class samples (repulsion property) in the latent space. Nearest neighbor based methods try to amplify the diversity in the attraction process. The source of this diversity is the process of sampling the positives in the form of nearest neighbors and not augmented views. However, this amplification of diversity introduces a trade-off between the attraction and repulsion properties. Because, if the quality of the support set is low or the chosen nearest neighbors are incorrect (section \ref{subsec:pnnclr}), $f$ may negatively impact the main objective by reducing the intended attraction and repulsion properties. To improve this trade-off, independent of the support set quality, we hypothesize modifying $f$ such that the diversity in the attraction process is amplified favorably, i.e., the positive samples retain the semantic variations and, at the same time, are not unfavorably distinct from the anchor point. We accomplish this by reducing the magnitude of the resultant vector in the direction of the nearest neighbor by a factor. Although, this controls the diversification in the attraction process, however, it reduces the semantic quality. To avoid this, a stochastic prior is imposed during the sampling of positives. This allows the expansion of uncertainty to increase the semantic information. As a consequence of these adaptations, the positives become more semantically diverse and related to the anchor point, thereby, helping in learning better representations by the model. We also found that by employing a smooth-weight-updation approach, the effects of uncertainty, introduced by nearest neighbor based learning, can be stabilized to a significant extent.

We tested the proposed adaptations and found that they significantly improve the performance of the proposed method over our baseline, NNCLR \cite{nnclr}, on image recognition tasks. Following this, we also tested it for the medical datasets and found a favorable performance. Following are the main contributions of this work:
\begin{itemize}
    \item We studied the suitability of nearest neighbor based semantic information enrichment in CL, and introduced stochastic pseudo nearest neighbors (pNN) to control the quality of the positive samples in CL based SSL methods. To the best of our knowledge, we are first to introduce pNN approach.
    \item We showed that a smooth-weight-updation approach in NN based CL methods is highly useful in controlling the uncertainty in sampling. Using this, we propose our CL base method called pNNCLR, which significantly improves over our baseline NNCLR \cite{nnclr}.
    \item We performed many experiments and ablation studies to empirically verify the superiority of our contribution through medical as well as non-medical datasets.
\end{itemize}

The remainder of the paper is organized as follows. Section \ref{sec:related work} presents the related literature on representation learning methods. Section \ref{sec:method} presents the details of our baseline and the proposed method. Section \ref{sec:experiments} presents the implementation details, evaluation, and comparison of the results on different datasets. After this, the conclusion and appendix are presented.

\section{Related Work}
\label{sec:related work}
Representation learning methods have underpinned some of the recent highly successful pretrained networks and amazing feats of AI --- GPT 3 \cite{gpt3brown2020language}, CLIP \cite{clipradford2021learning}, SAM \cite{samkirillov2023segment}, ChatGPT \cite{chatgptliu2023summary}, SEER \cite{seergoyal2021self}. These representation learning methods are mainly trained in a self-supervised manner. A couple of years earlier, self-supervised approaches were dominated by non-contrastive methods or by pretext task based methods (Context prediction \cite{doersch2015unsupervised}, Jigsaw \cite{noroozi2016unsupervised}, RotNet \cite{gidaris2018unsupervised}). However, the trend is shifting as the current best self-supervised methods are all based on some form of contrastive learning approach (SBCL \cite{hou2023subclass}, DINO \cite{dino}). Our baseline, NNCLR \cite{nnclr}, is one such contrastive learning based method that employs nearest neighbor sampling. In this literature review, we cover the developments in SSL from the perspective of both --- non-contrastive and contrastive --- self-supervised methods. Table \ref{tab:literature review} provides a consolidated view of the literature.

\begin{table}[t!]
    \caption{A consolidated literature review is presented for both --- contrastive and non-contrastive SSL methods. Note that most of the backbones are CNN based.}
    \centering
    \resizebox{1\linewidth}{!}{
		\begin{tabular}{lccccl}
		\toprule
		Author         & Year & Method                & Contrastive & Backbone            & Approach / Pretext-task                                                                                                                                                                           \\ \midrule
		Doersch et al. \cite{doersch2015unsupervised} & 2015 & SpatialContext        & \ding{55}          & VGG                 & \begin{tabular}[l]{@{}l@{}}Predicting spatial context of a patch in relation\\ to another patch in a spatially consistent\\ array of nine patches.\end{tabular}                                      \\
		Zhang et al. \cite{zhang2016colorful}   & 2016 & CCEncoder             & \ding{55}          & In-house, VGG       & \begin{tabular}[l]{@{}l@{}}Cross channel prediction using\\ auto-encoder network.\end{tabular}                                                                                                                                               \\
		Pathak et al. \cite{pathak2016context}  & 2016 & ContextEncoder        & \ding{55}          & AlexNet             & \begin{tabular}[l]{@{}l@{}}Inpainting of missing patch using context\\ auto-encoders with channelwise\\ fully-connected layers.\end{tabular}                                                         \\
		Misra et al. \cite{misra2016shuffle}   & 2016 & ShuffleAndLearn       & \ding{55}          & SiameseAlexNet      & \begin{tabular}[l]{@{}l@{}}Ordering of frames with sequence\\ binary verification.\end{tabular}                                                                                                                                              \\
		Noroozi et al. \cite{noroozi2016unsupervised} & 2016 & ContextFreeNetwork    & \ding{55}          & SiameseAlexNet      & \begin{tabular}[l]{@{}l@{}}Rearrangement of shuffled Jigsaw puzzle like\\ sub-images.\end{tabular}                                                                                                                                           \\
		Zhang et al. \cite{zhang2017split}   & 2017 & SplitBrainAutoEncoder & \ding{55}          & Channelwise AlexNet & \begin{tabular}[l]{@{}l@{}}Correct prediction of rearranged incomplete\\ input channels.\end{tabular}                                                                                                                                        \\
		Gidaris et al. \cite{gidaris2018unsupervised} & 2018 & RotNet                & \ding{55}          & AlexNet             & \begin{tabular}[l]{@{}l@{}}Correct prediction of rotated images.\end{tabular}                                                                                                                                                              \\
		Oord et al. \cite{infonce}    & 2018 & CPC                   & \ding{51}         & ResNet-v2-101       & \begin{tabular}[l]{@{}l@{}}Using contrastive predictive coding in PixelCNN\\ auto-regressive recurrent neural networks\\ for prediction of output embedding vectors.\end{tabular}                  \\
		Caron et al. \cite{caron2020unsupervised}   & 2020 & SwAV                  & \ding{51}         & ResNet-50           & \begin{tabular}[l]{@{}l@{}}Instead of pairwise contrastive loss, an online\\ clustering of multiple views of same image is\\ performed to learn the features\\ and cluster assignments.\end{tabular} \\
		Chen et al. \cite{chen2020generative}    & 2020 & iGPT-XL               & \ding{55}          & GPT-2 BERT          & \begin{tabular}[l]{@{}l@{}}Auto-regressive prediction of pixels\\ using transformers.\end{tabular}                                                                                                                                           \\
		Chen et al. \cite{simclr}    & 2020 & SimCLR                & \ding{51}         & ResNet-50           & \begin{tabular}[l]{@{}l@{}}A simple approach based on contrastive loss,\\ non-linearity layer, augmentations and\\ large batch sizes.\end{tabular}                                                    \\
		He et al. \cite{moco}      & 2020 & MoCo                  & \ding{51}         & ResNet-50           & \begin{tabular}[l]{@{}l@{}}An online dictionary approach for contrastive\\ learning using memory bank and\\ momentum contrast.\end{tabular}                                                          \\
		Grill et al. \cite{byol}   & 2020 & BYOL                  & \ding{51}         & ResNet-50           & \begin{tabular}[l]{@{}l@{}}Does not use negative pairs for contrastive loss\\ while using momentum contrast.\end{tabular}                                                                          \\
		Chen et al. \cite{simsiam}    & 2021 & SimSiam               & \ding{51}         & ResNet-50           & \begin{tabular}[l]{@{}l@{}}Uses a simple Siamese network without negative\\ pairs, large batch size, momentum contrast.\end{tabular}                                                                 \\
		Caron et al. \cite{dino}   & 2021 & DINO                  & \ding{51}         & ViT-S/16            & \begin{tabular}[l]{@{}l@{}}Contrastive learning on vision transformers\\ using a codistillation approach.\end{tabular}                                                                             \\
		Goyal et al. \cite{seergoyal2021self}   & 2021 & SEER                  & \ding{51}         & RegNet-Y            & \begin{tabular}[l]{@{}l@{}}SwAV method is used to train large SSL model\\ on very large dataset in the wild.\end{tabular}                                                                          \\
		Dwibedi et al. \cite{nnclr} & 2021 & NNCLR                 & \ding{51}         & ResNet-50           & \begin{tabular}[l]{@{}l@{}}Uses a nearest neighbor approach to increase the\\ semantic variation during learning.\end{tabular}                                                                     \\
		Xie et al. \cite{xie2022simmim}     & 2022 & SimMIM                & \ding{55}          & ViT-B               & \begin{tabular}[l]{@{}l@{}}Correct prediction of patch level masked\\ images using transformers.\end{tabular}                                                                                                                                \\
		Yeh et al. \cite{dclr}     & 2022 & Decoupled CLR         & \ding{51}         & ResNet-50           & \begin{tabular}[l]{@{}l@{}}Remove the positive term from the denominator\\ of the InfoNCE loss to reduce the\\ positive-negative-coupling.\end{tabular}                                              \\
		Zhang et al. \cite{zhang2023patch}   & 2023 & ADCLR                 & \ding{51}         & ViT-S/16            & \begin{tabular}[l]{@{}l@{}}Transformer based approach for dense contrastive\\ learning by balancing global and\\ patch-level losses.\end{tabular}                                                   \\
		Hou et al. \cite{hou2023subclass}     & 2023 & SBCL                  & \ding{51}         & ResNet-50           & \begin{tabular}[l]{@{}l@{}}A hierarchical online clustering like SwAV to\\ balance the emphasis between head class\\ and long-tailed class.\end{tabular}                                       \\ \bottomrule
		\end{tabular}
    }
    \label{tab:literature review}
\end{table}

\subsection{Non-contrastive SSL methods}
In the context of self-supervised learning, a \textit{pretext task} refers to a puzzle or sub-task that is solved by the SSL model. The objective is to learn the underlying structure of the data by deriving a supervision signal from the sub-task in an unsupervised manner, i.e., without relying on any labeled data \cite{doersch2015unsupervised}.
Doersch et al. explored spatial context as a supervision signal for training visual representations \cite{doersch2015unsupervised}. Their approach involved dividing a region in image into 9 patches, then sampling pairs from these patches and training the model to predict the relative position of a patch given another patch from the pair. They achieved unsupervised object discovery and improved performance on object detection tasks.
Zhang et al. used cross channel color space prediction as a pretext task \cite{zhang2016colorful,zhang2017split}. They used a CNN to predict \textit{ab} color space from \textit{L} channel of the CIE \textit{Lab}* color space. It was found that colorization could be a useful option for learning representations for vision tasks.
Pathak et al. also used context information for their inpainting pretext task by forcing their context encoding CNN to predict the missing region in an input image \cite{pathak2016context}. Although predicting the entire region is an under constrained task, however, their approach produced strong image representations.
Another line of work (Shuffle learn \cite{misra2016shuffle}, Sequence sorting \cite{lee2017unsupervised}, Sustained order verification \cite{buckchash2019sustained,buckchash2020dutrinet}, Odd one out \cite{fernando2017self}), used frame order prediction as the pretext task. These models learned meaningful image representations for vision tasks. However, it was not as strong as the representations learned by other pretext tasks, like context prediction or spatial rearrangement.
Noroozi et al. proposed an even more challenging pretext task of sorting all nine pieces of a Jigsaw puzzle \cite{noroozi2016unsupervised}. They also suggested several shortcut prevention approaches as they emphasized --- ``A good self-supervised task is neither simple nor ambiguous."
Gidaris et al. proposed a simple rotation prediction as the pretext task for CNNs \cite{gidaris2018unsupervised}. The objective was to predict the angle of rotation from {0, 90, 180, 270} degrees. Rotation turned out to be a simple yet powerful SSL strategy since it does not leave any easily detectable low-level visual shortcut for trivial feature learning.
Chen et al. adapted a GPT-2 scale transformer model from Masked Language Modeling (MLM) to Masked Image Modeling (MIM) on pixels of down-scaled images \cite{chen2020generative}. Objective of the pretext task was to auto-regressively predict the masked pixels in the transformer output in a BERT-like sense. 
Following the same line, Zhou et al. introduced an online visual tokenizer for MIM \cite{zhou2021ibot}. They showed that better semantics could be learned by simultaneously training the tokenizer with the MIM transformer through knowledge distillation.
Xie et al. simplified the previous transformer based masked prediction pretext task methods by dropping blockwise masking and tokenization \cite{xie2022simmim}. Their model achieved competitive results with just linear probing.

\subsection{Contrastive SSL methods}
Although, pretext based methods achieved good representations, however, Misra et al. showed that they all followed a covariant style of modeling \cite{misra2020self}. Misra et al. advocated the superiority of an invariant style of modeling over a covariant. Their work sits between non-contrastive and contrastive SSL methods. The main contribution is the noise contrastive estimation formulation which involves the generation of positive and negative pairs using the Jigsaw objective \cite{noroozi2016unsupervised}. Contrastive SSL methods differ in their approach by including the pretext task in the model architecture itself in the form of augmentations. Additionally, the model objective changes from equivariance (where the model tries to adjust itself according to the variation in input) to invariance (where the model ignores the changes in the input in order to become agnostic to those transformations). To be specific, the SSL supervision signal is derived by enforcing the equivalence of multiple views of the same input image \cite{simclr}.
One such initial work by Oord et al. proposed Contrastive Predictive Coding (CPC) for unsupervised representation learning \cite{infonce}. They applied noise contrastive loss (NCL) over future predictions in latent space of auto-regressive models for speech, text, and images. An important aspect of their NCL formulation was inclusion of negative samples.
Later it was picked up and improved by Chen et al. \cite{simclr}. They proposed a simple contrastive learning approach in which they emphasized on compositions in data augmentation, role of non-linear transformations in top layers, and larger batch size.
He et al. proposed the idea of dictionary look-up by maintaining a dictionary of encoded keys as negatives for contrastive learning \cite{moco}. This allowed keeping a larger set than the batch size as negatives. Similar to He et al., Girll et al. proposed online and target network based contrastive learning where the target network avoids direct gradient flow and takes updates from the online network \cite{byol}. They also showed that the setup does not require negative samples for training the network.
To reduce the overall computation in contrastive SSL methods, Caron et al. proposed to avoid pairwise comparisons by employing online clustering of the representations and by enforcing consistency in cluster assignments of representations corresponding to different views of the same sample \cite{caron2020unsupervised}.
Chen et al. refined the contrastive SSL aspects of previous works \cite{simsiam}. They trained a Siamese network with contrastive loss but without negative sample pairs and without large batch sizes. They avoided trivial solutions and attained competitive performance by avoiding gradient propagation in one of the branches of the Siamese network.
Caron et al. proposed a self-supervised knowledge distillation approach called DINO \cite{dino}. They showed that vision transformers learn better semantic segmentation and \textit{k}-NN features than CNNs.
Goyal et al. validated the contemporary contrastive SSL approaches in the wild by training with one billion random images \cite{seergoyal2021self}.
Yeh et al. proposed decoupling the positive samples from the denominator of the InfoNCE contrastive loss to remove the negative-positive-coupling effect in contrastive SSL methods \cite{dclr}.
Zhang et al. introduced query patches for contrasting in addition to global contrasting \cite{zhang2023patch}.
Nearest neighbor based methods like \cite{nnclr,hou2023subclass} emphasized increasing semantic variation by sampling the nearest neighbors in the latent space. The proposed work is similar in spirit to the nearest neighbor sampling based works.

\section{Method}
\label{sec:method}
In this section, we first formalize the representation learning problem. After this, the NNCLR approach is described, and finally, we present the details of the proposed method.

\subsection{Problem}
Representation learning aims to learn a model which can map its inputs to corresponding vectors in such a way that for any two closely related inputs, their vectorial representations are also close, and far away for any two unrelated or distinctly related inputs. For a given dataset $\mathcal{X}$ having images $x_i | i \in [1, N]$, we wish to learn a homomorphism, model $\mathcal{M}_\theta$ parameterized by $\theta$, such that for any three inputs $x_i$, $x_j$ and $x_k$, $\mathcal{M}_\theta$ gives three corresponding vectors $\bm{u}_i$, $\bm{u}_j$, and $\bm{u}_k$ respectively, such that the following condition holds:

\begin{equation}
    d_x(x_i, x_j) \star d_x(x_i, x_k) \Leftrightarrow d_v(\bm{u}_i, \bm{u}_j) \star d_v(\bm{u}_i, \bm{u}_k),
\end{equation}

where, $d_x$ and $d_v$ represent the distance function in image and vector spaces respectively, and $\star$ is any relational operator like $\ll$.

\subsection{NNCLR}
\begin{figure}[t]
    \centering
    \begin{subfigure}[b]{0.3\textwidth}
        \centering
        \includegraphics[width=\textwidth]{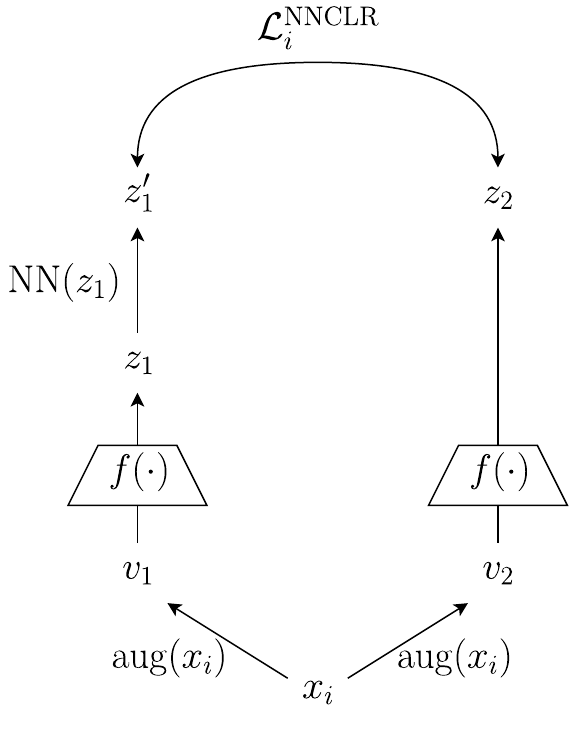}
        \caption{NNCLR}
        \label{subfig:nnclr}
    \end{subfigure}
    \hspace{2em}
    \begin{subfigure}[b]{0.3\textwidth}
        \centering
        \includegraphics[width=\textwidth]{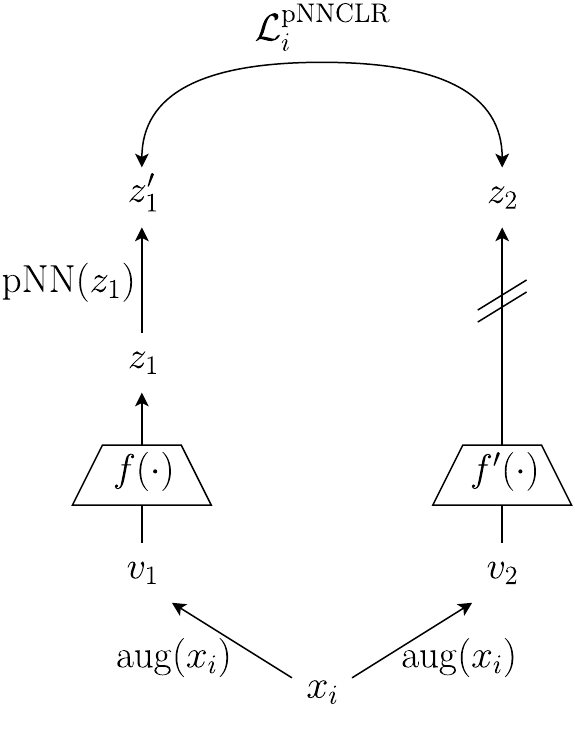}
        \caption{pNNCLR}
        \label{subfig:pnnclr}
    \end{subfigure}
    \caption{Model diagrams of NNCLR and pNNCLR methods. The cross sign in pNNCLR denotes stop-gradient operation. $f(\cdot)$ is the backbone encoder network. ${\rm aug}(x_i)$ is a random transformation function that generates a new view for $x_i$. ${\rm pNN}(\cdot)$ is the proposed pseudo nearest neighbor sampling function.}
    \label{fig:model}
\end{figure}

Contrastive learning methods like SimCLR \cite{simclr} or BYOL \cite{byol} train by generating two augmented views $v_1, v_2$ for the same input $x_i$. During the loss calculation, embeddings corresponding to $v_1, v_2$ are treated as positives, whereas embeddings corresponding to all other $x_j | j \neq i$ are treated as negatives to $v_1, v_2$. A variant of InforNCE loss \cite{infonce} like

\begin{equation}
    \label{eq:loss simclr}
    \mathcal{L}^{SimCLR}_{i} = - \log \frac{\exp{(z_i \cdot z^{+}_i / \tau)}}{\sum\limits_{k=1}^{n}\exp{(z_i \cdot z^{+}_k / \tau)}},
\end{equation}

is used, where $z_i$ is the embedding or the vector corresponding to the view $v_1$, $z^{+}_i$ is the positive pair of $z_i$. The set, $z^{+}_k | k\in [1,n]$, denotes all embeddings in the mini-batch (with size $n$), including the positive $z^+_i$ and negatives $z^-_k | k\neq i$. $\tau$ denotes the softmax temperature. The operation, $\bm{u} \cdot \bm{v}$ in Eq. \eqref{eq:loss simclr}, represents a similarity function, generally a dot product of the normalized vectors $\bm{u}$, $\bm{v}$ or their cosine similarity.
NNCLR improves this approach by replacing $z_i$ with its nearest neighbor, ${\rm NN}(z_i)$, as shown in Fig. \ref{subfig:nnclr}. The nearest neighbor is found from a support set $Q$, which is maintained by inserting the current batch items and removing the oldest batch items from it in a first-in-first-out manner for every training iteration. Using ${\rm NN}(z_i)$, NNCLR loss function for $x_i \in {\rm batch}\{ x_k | 1\leq k \leq n\}$ becomes:

\begin{equation}
    \label{eq:loss nnclr}
    \mathcal{L}^{NNCLR}_{i} = - \log \frac{\exp{({\rm NN}(z_i) \cdot z^{+}_i / \tau)}}{\sum\limits_{k=1}^{n}\exp{({\rm NN}(z_i) \cdot z^{+}_k / \tau)}}.
\end{equation}

\subsection{pNNCLR}
\label{subsec:pnnclr}
\begin{figure}[t]
    \centering
    \begin{subfigure}[b]{0.46\textwidth}
        \centering
        \includegraphics[width=\textwidth]{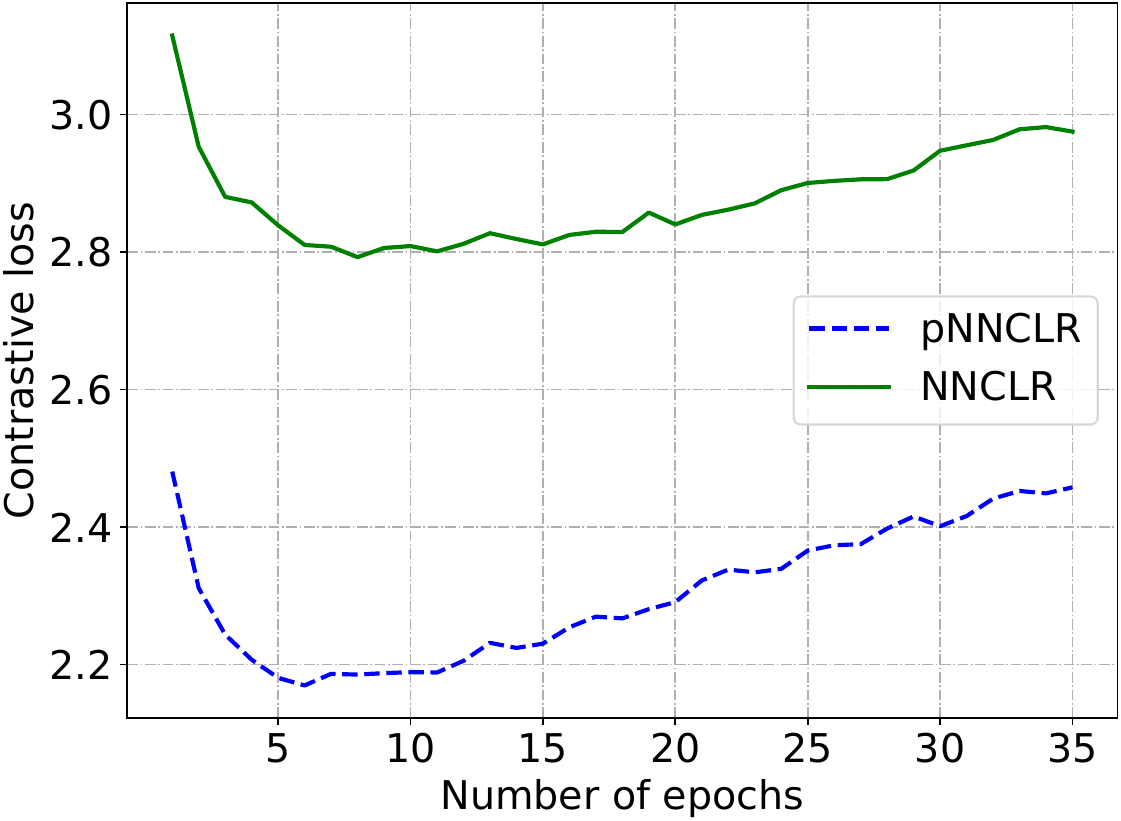}
        \caption{Tiny-imagenet}
        \label{subfig:tinyimagenetlossplot}
    \end{subfigure}
    \hfill
    \begin{subfigure}[b]{0.46\textwidth}
        \centering
        \includegraphics[width=\textwidth]{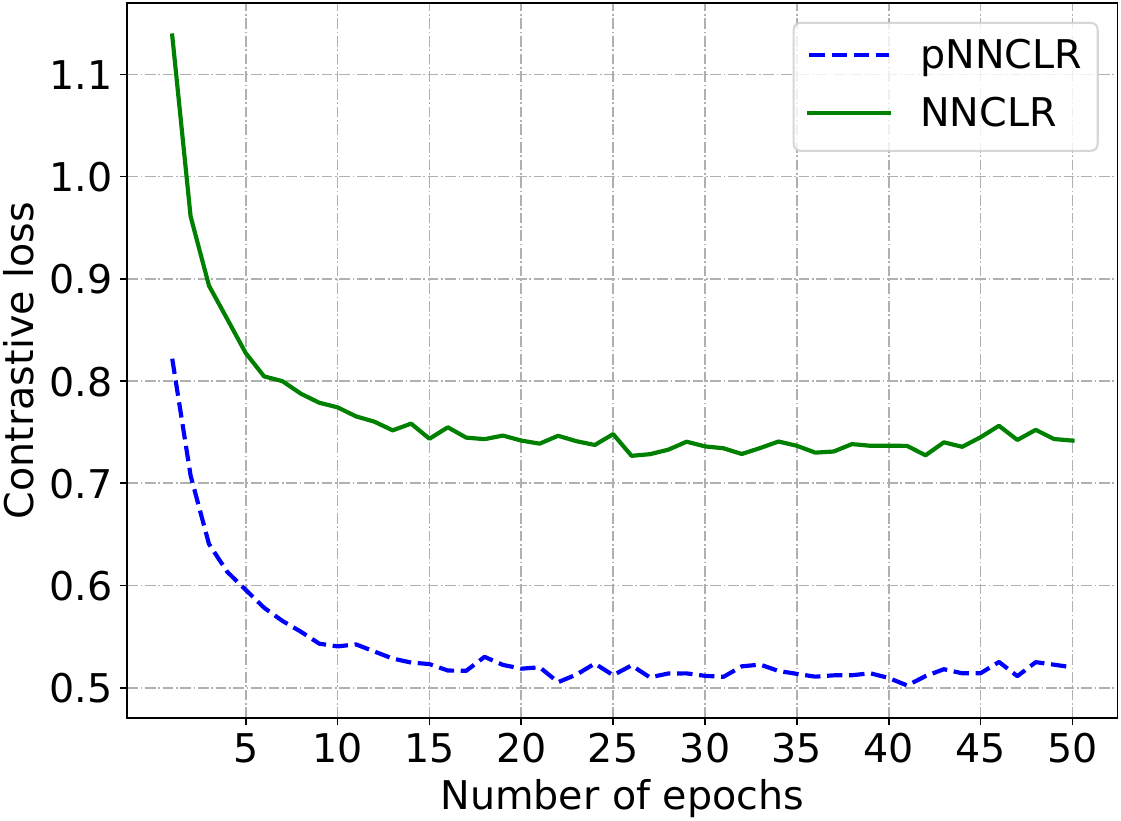}
        \caption{STL-10}
        \label{subfig:stl10lossplot}
    \end{subfigure}
    \caption{Loss plots of NNCLR and pNNCLR methods on Tiny-imagenet and STL-10 datasets. Note,  NNCLR incurs a higher loss right from the beginning of training.}
    \label{fig:lossplots}
\end{figure}

Although the intuition of semantic variability behind NNCLR has shown promising results compared to other recent developments in contrastive learning methods \cite{nnclr}, however, NNCLR achieves this at a cost, since the probability of finding a hard nearest neighbor, from the support set $Q$, belonging to the same class is quite low ($\sim$ 50\%) at the beginning of training (details in \ref{app:same class nn probability}).
I.e., if ${\rm class}(\cdot)$ denotes the class membership function, the approximate (since it depends on the support set size) maximal probability,  $P[{\rm class}({\rm NN}(z_i)) = {\rm class}(z_i)]$, that the nearest neighbor belongs to the same class as $z_i$ or $x_i$ is around 50\% (details in \ref{app:same class nn probability}), which means there is a 50\% chance that the nearest neighbor is from a different class. This reduces the inter-class variation of the representations, leading to a decline in performance.
This was also seen in the NNCLR method's loss plots (Fig. \ref{fig:lossplots}). NNCLR incurs a higher loss in the beginning of training.

If we carefully investigate the intuition behind NNCLR, we will find that it is trying to increase the semantic variability between the two views $v_1$, $v_2$. However, by doing so, it is also dispersing the intra-class representations to have a larger mean deviation. These two ideas are inversely related. To overcome this trade-off, this work proposes to use soft or pseudo nearest neighbor function, ${\rm pNN}(\cdot)$, in place of ${\rm NN}(\cdot)$, to perform better irrespective of the probability $P[{\rm class}({\rm NN}(z_i)) = {\rm class}(z_i)]$. The proposed method is called, pNNCLR, pseudo/probabilistic nearest neighbor CLR (Fig. \ref{subfig:pnnclr}). Function ${\rm pNN}(\cdot)$, works by sampling a point $z^{\prime\prime}_i$ in the direction of the vector $\overrightarrow{z_i\; {\rm NN}(z_i)}$ such that the resultant vector $\overrightarrow{z_i\; z^{\prime\prime}_i}$ has a shorter magnitude than $\overrightarrow{z_i\; {\rm NN}(z_i)}$. This shortness is controlled by a scalar hyperparameter $\alpha \in (0,1)$ as:

\begin{equation}
    \label{eq:pseudo nn sampling alpha}
    z^{\prime\prime}_i \leftarrow z_i + (1-\alpha)({\rm pNN}(z_i) - z_i).
\end{equation}

While the probability $P[{\rm class}({\rm NN}(z_i)) = {\rm class}(z_i)]$ improves by using $z^{\prime\prime}_i$ over $z_i$, some semantic variability is lost. To dilute this effect, we found that we can stochastically resample in the vicinity of $z^{\prime\prime}_i$. This is done by using a Gaussian prior with mean $z^{\prime\prime}_i$ and standard deviation which is a fraction, $\beta$, of $\lVert \overrightarrow{z_i\; z^{\prime\prime}_i} \rVert$, where $\lVert \cdot \rVert$ denotes magnitude of a vector. This is shown in Eq. \eqref{eq:pseudo nn sampling beta}.

\begin{equation}
    \label{eq:pseudo nn sampling beta}
    z^{\prime}_i \sim \mathcal{N}(z^{\prime\prime}_i, \beta \lVert \overrightarrow{z_i\; z^{\prime\prime}_i} \rVert),
\end{equation}

where, $\mathcal{N}$ stands for a normal distribution, and $\beta \in (0,1)$ is a scalar constant. Due to higher uncertainty in NN based approach, we slow down the weight updation process of the encoder network $f^\prime(\cdot)$, shown in Fig. \ref{subfig:pnnclr}, by stopping the gradient flow in non ${\rm pNN}(\cdot)$ branch (also called the target branch \cite{byol}). Providing a smoother updation of weights by using following:

\begin{equation}
    \label{eq:target branch updation}
    \theta_{f^\prime} \leftarrow \lambda \theta_{f^\prime} + (1-\lambda)\theta_{f},
\end{equation}

where, $\theta$ stands for network parameters, $f$ is the online network, $f^\prime$ is the target network, $\lambda \in (0,1)$ is a constant that controls the effect of $f$ over $f^\prime$. By replacing the nearest neighbor function in Eq. \ref{eq:loss nnclr}, we obtain the loss for pNNCLR, as:

\begin{equation}
    \label{eq:loss pnnclr}
    \mathcal{L}^{pNNCLR}_{i} = - \log \frac{\exp{({\rm pNN}(z_i) \cdot z^{+}_i / \tau)}}{\sum\limits_{k=1}^{n}\exp{({\rm pNN}(z_i) \cdot z^{+}_k / \tau)}},
\end{equation}

using this, the loss for the entire mini-batch, $b$, of size $N_b$ becomes:

\begin{equation}
    \label{eq:loss pnnclr mini-batch}
    \mathcal{L}^{pNNCLR}_b = \frac{1}{N_b} \sum\limits_{i=1}^{N_b} \mathcal{L}^{pNNCLR}_{i}.
\end{equation}

The total loss, $\mathcal{L}^{pNNCLR}$, is a symmetrized loss obtained by swaping the views $v_1$, $v_2$ in Eq. \eqref{eq:loss pnnclr mini-batch}, as:

\begin{equation}
    \label{eq:loss pnnclr total}
    \mathcal{L}^{pNNCLR} = \mathcal{L}^{pNNCLR}_b(v_1, v_2) + \mathcal{L}^{pNNCLR}_b(v_2, v_1).
\end{equation}

\section{Experiments}
\label{sec:experiments}
This section first describes the experimental arrangement. Next, the implementation and dataset related details are presented. After this, the results of the proposed pNNCLR approach are compared with the recent methods for SSL for the linear evaluation task. Towards the end, some ablations, discussion on results, and future directions are presented.

\subsection{Implementation details}
\begin{figure}[t]
	\centering
	\includegraphics[width=\linewidth]{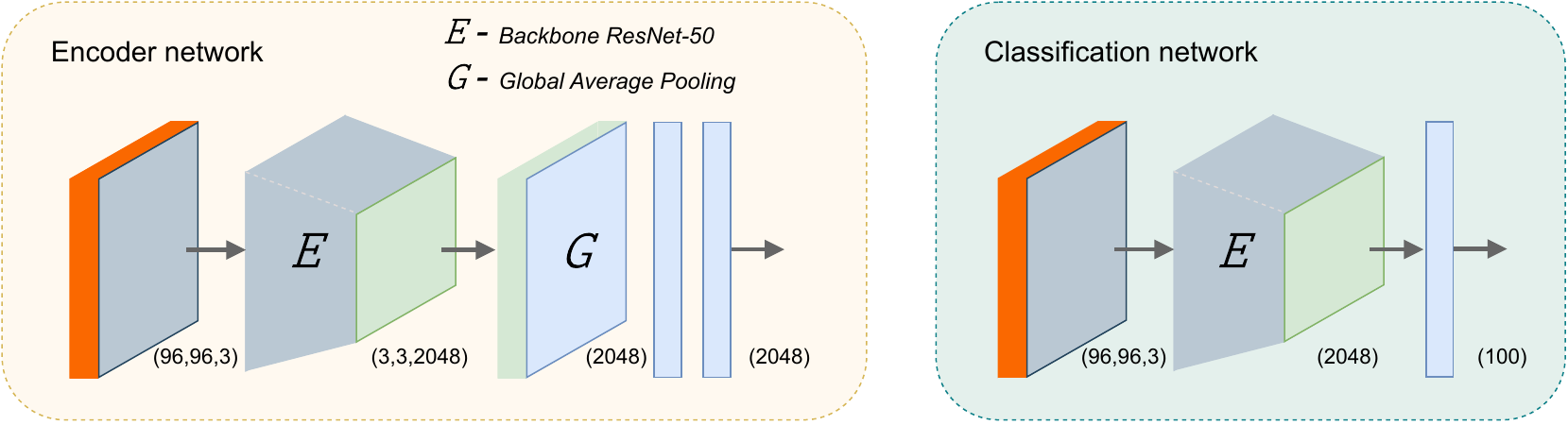}
	\caption{Proposed pNNCLR architecture details considering Cifar-100 as the downstream task. Left, architecture for contrastive SSL training. Right, linear probing adaptation on Cifar-100 dataset.}
	\label{fig:pnnclr architecture stl-10}
\end{figure}

We have used the batch size of 64, and 10000 as the size of the support set. The embedding size was kept at 2048. The optimizer was Adam, and the learning  rate was set to 0.001. The images for every dataset were resized to 96$\times$96. ResNet-50 \cite{he2016deep} was used as the base encoder network. The final prediction layer of ResNet was removed, and a Global average pooling was used for flattening, followed by two dense layers having 2048 nodes, and a batch norm was present between these two dense layers as shown in Fig. \ref{fig:pnnclr architecture stl-10}, encoder network. During testing, the encoder was frozen after the fine-tuning, and an additional dense layer having the softmax activation was used for classification (linear probing), as shown in Fig. \ref{fig:pnnclr architecture stl-10}, classification network. Five non-medical datasets (STL-10, Cifar-10,100, Tiny-imagenet, Pascal-VOC) and three medical datasets (Blood-MNIST, PCAM, Path-MNIST) were used for evaluation and comparison purposes. Details of these datasets and their splitting strategy for training and testing purposes is provided in the next section. Top-1 accuracy was the metric used for all our experiments unless stated otherwise.

\subsection{Datasets}
\paragraph{STL-10} It is a standard dataset derived from Imagenet \cite{imagenet} for developing self-supervised learning algorithms. It has 100000 unlabeled and 13000 labeled images from 10 classes (like bird, cat, truck) \cite{stl10}. All models reported here, were trained for 100 epochs on this dataset.

\paragraph{Cifar-10 and Cifar-100} Each of these datasets consists of 60000 images \cite{cifar}. Cifar-10 consists of 10 classes, whereas Cifar-100 consists of 100 classes. General class labels are bird, dog, ship, horse, truck, etc.

\paragraph{Tiny-imagenet} This dataset contains 120000 images from 200 classes \cite{tinyimagenet}.

\paragraph{Pascal-VOC} This dataset contains 20 classes like vehicles, airplanes, animals, etc. It contains approximately 3000 images.

\begin{table}[t]
    \caption{Results are shown for Top-1 accuracy on the non-medical datasets on the image recognition task. For each dataset, \textbf{best} method is marked in bold, \underline{second best} is underlined.}
    \centering
    \resizebox{1\linewidth}{!}{
        \begin{tabular}{lcccccc}
        \toprule
        Method                   & STL-10 & Cifar-10 & Cifar-100 & Tiny-imagenet & Pascal-VOC & Mean (Top-1 acc.) \\ \midrule
        Baseline (NNCLR \cite{nnclr})         & 0.7548 & 0.9441   & 0.7763    & 0.3929        & -          & 0.7170            \\
        MoCo v2 \cite{moco}                  & \underline{0.8355} & 0.9411   & 0.7804    & 0.4651        & \underline{0.4900}     & \underline{0.7555}            \\
        BYOL \cite{byol}                     & 0.8044 & 0.9456   & \underline{0.7882}    & 0.4577        & -          & 0.7489            \\
        SimCLR \cite{simclr}                   & 0.7974 & 0.9428   & 0.7812    & \underline{0.4660}        & -          & 0.7468            \\
        SimSiam \cite{simsiam}                  & 0.8067 & 0.9418   & 0.7811    & 0.3284        & -          & 0.7145            \\
        DINO \cite{dino}                     & 0.8200 & 0.9459   & 0.7809    & 0.3161        & -          & 0.7157            \\
        Decoupled CLR \cite{dclr}            & 0.8343 & \underline{0.9498}   & 0.7812    & 0.4245        & -          & 0.7474            \\
        Proposed method (pNNCLR) & \textbf{0.8413} & \textbf{0.9582}   & \textbf{0.7885}    & \textbf{0.4856}        & \textbf{0.5066}     & \textbf{0.7684}            \\ \bottomrule
        \end{tabular}
    }
    \label{tab:results all non-medical}
\end{table}

\paragraph{Blood-MNIST and Path-MNIST} Both of these datasets belong to a large collection of biomedical images \cite{medmnist}. Blood-MNIST has approximately 17000 images belonging to 8 classes from blood cell microscopy. Path-MNIST has 9 classes having approximately 100000 images of Colon pathology.

\paragraph{PCAM or PatchCamelyon} It is a binary image classification dataset \cite{pcam} having approximately 327000 images extracted from histopathologic scans of lymph node sections to indicate the presence of metastatic tissue. 

Features learned from STL-10 were used to apply transfer learning to other datasets. Approximately $\sim$ 2\% of the images were used for applying transfer learning using a linear layer on the pretrained encoder model.

\subsection{Results}
To evaluate the performance of the proposed method, it is compared with very competitive recent benchmark self-supervised learning works. The results are presented in Table \ref{tab:results all non-medical} and \ref{tab:results all medical} for non-medical and medical types of datasets, respectively. Our baseline work is the NNCLR method \cite{nnclr}, which was published in ICCV 2021. MoCo \cite{moco} and BYOL \cite{byol} are momentum contrast based approaches for contrastive learning, and appeared in CVPR 2020 and NeurIPS 2020 respectively. SimCLR \cite{simclr} is the baseline approach of NNCLR, and was published in PMLR 2020. SimSiam \cite{simsiam} showed that even without using the negative samples, good performance could be achieved in SSL. It was published in CVPR 2021. DINO \cite{dino} is a vision transformer based method and was published in ICCV 2021. Decoupled CLR \cite{dclr} removed the positives from the denominator of the InfoNCE loss \cite{infonce} and proposed a simple CL method which was published in ECCV 2022.

\begin{table}[t]
    \caption{Results are shown for Top-1 accuracy on the medical datasets on the image recognition task. For each dataset, \textbf{best} method is marked in bold, \underline{second best} is underlined.}
    \centering
    \resizebox{.9\linewidth}{!}{
        \begin{tabular}{lcccc}
        \toprule
        Method                   & Blood-MNIST & PCAM   & Path-MNIST & Mean (Top-1 acc.) \\ \midrule
        Baseline (NNCLR \cite{nnclr})         & 0.7969      & 0.8849 & 0.8292     & 0.8370            \\
        MoCo v2 \cite{moco}                  & \textbf{0.8906}      & \textbf{0.9180} & 0.8562     & \textbf{0.8882}            \\
        BYOL \cite{byol}                     & 0.8516      & 0.8868 & 0.8365     & 0.8583            \\
        SimCLR \cite{simclr}                   & 0.8594      & 0.8951 & 0.8552     & 0.8699            \\
        SimSiam \cite{simsiam}                  & 0.8594      & 0.8397 & 0.7917     & 0.8302            \\
        DINO \cite{dino}                     & 0.7500      & 0.7824 & 0.7698     & 0.7674            \\
        Decoupled CLR \cite{dclr}            & 0.8281      & 0.8884 & \underline{0.8615}     & 0.8593            \\
        Proposed method (pNNCLR) & \underline{0.8672}      & \underline{0.9025} & \textbf{0.8708}     & \underline{0.8801}            \\ \bottomrule
        \end{tabular}
    }
    \label{tab:results all medical}
\end{table}

\begin{figure}[t]
	\centering
	\includegraphics[width=.5\linewidth]{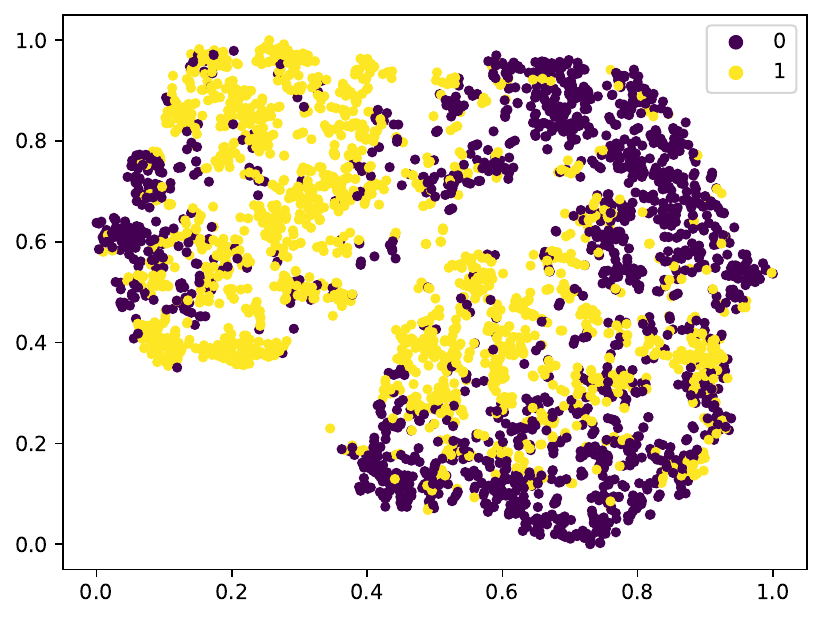}
	\caption{T-sne plot of the representations learned by the proposed pNNCLR method on the PatchCamelyon (PCAM) medical dataset.}
	\label{fig:tsne pcam}
\end{figure}

\begin{figure}[t]
    \centering
    \begin{subfigure}[b]{0.46\textwidth}
        \centering
        \includegraphics[width=\textwidth]{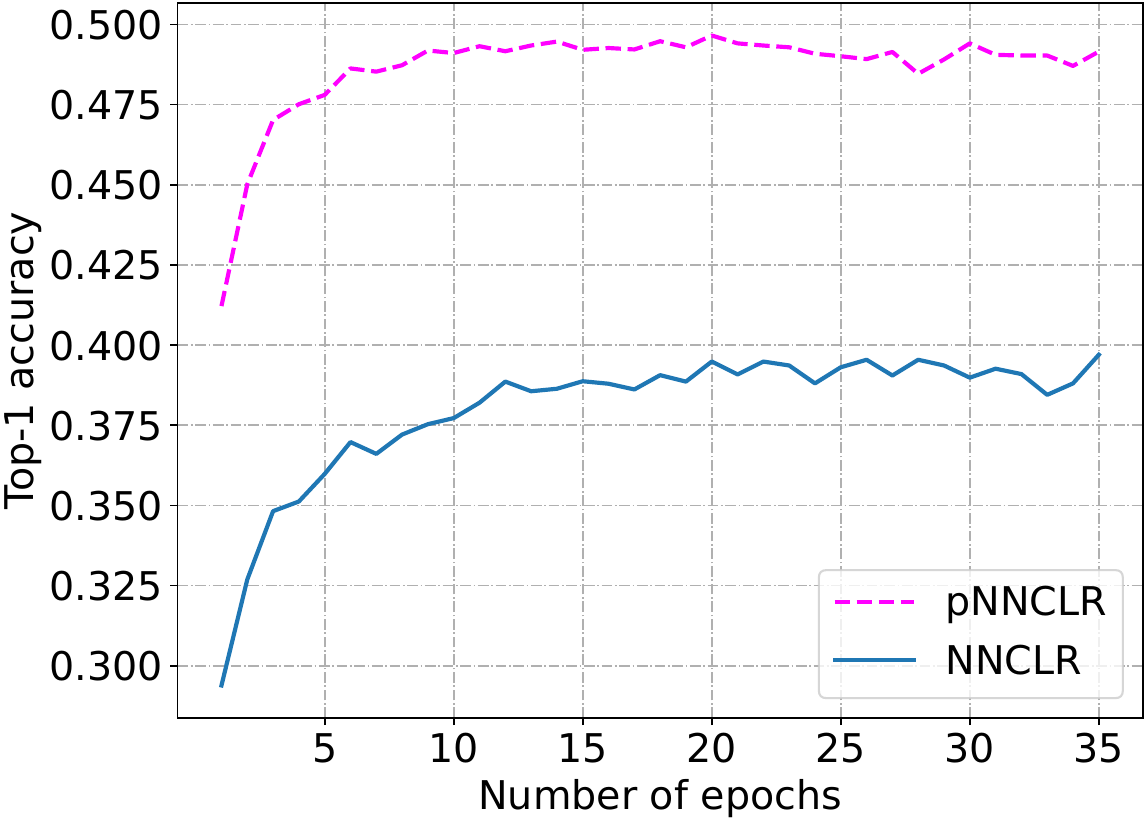}
        \caption{Tiny-imagenet}
        \label{subfig:tinyimagenetaccuracyplot}
    \end{subfigure}
    \hfill
    \begin{subfigure}[b]{0.46\textwidth}
        \centering
        \includegraphics[width=\textwidth]{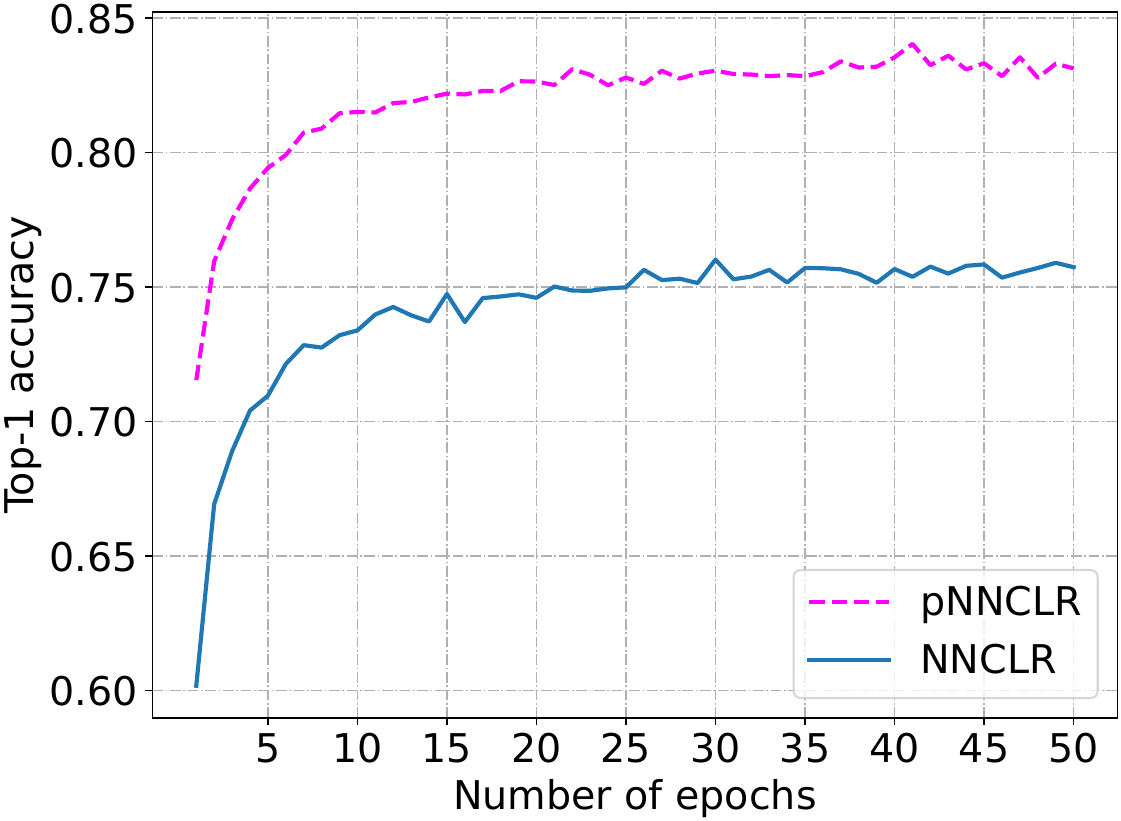}
        \caption{STL-10}
        \label{subfig:stl10accuracyplot}
    \end{subfigure}
    \caption{Top-1 accuracy plots of the baseline NNCLR and proposed pNNCLR methods on Tiny-imagenet and STL-10 datasets.}
    \label{fig:accuracyplots}
\end{figure}

Table \ref{tab:results all non-medical} presents the results for non-medical datasets. The proposed method, pNNCLR, achieved the highest Top-1 accuracy for each of the datasets among all methods. It surpassed the baseline, NNCLR, by a maximum $\sim$ 8\% on STL-10 and Tiny-imagenet, and by $\sim$ 4\% on average over all datasets. The second best performance was attained by MoCo \cite{moco}, which is $\sim$ 1\% less than pNNCLR, on average. Table \ref{tab:results all medical} presents the results for medical datasets. For the Blood-MNIST dataset, MoCo attained the best results with an accuracy of 89.06\% while the proposed pNNCLR method lagged behind by $\sim$ 2\%; however, it performed $\sim$ 7\% better than the baseline NNCLR. On the PCAM dataset, again the best performance was attained by MoCo while pNNCLR lagged behind by $\sim$ 1\%. On the Path-MNIST dataset, pNNCLR attained the best results while the second best performance was attained Decoupled CLR method. On average, pNNCLR lagged behind by the best performance by less than 1\% while surpassing the baseline NNCLR by $\sim$ 4\%. Figure \ref{fig:tsne pcam} presents a low dimensional view of the representations learned by the proposed pNNCLR method on the PCAM dataset for the binary classification of the presence of metastatic tissues in lymph node scans. Figure \ref{fig:accuracyplots} shows the accuracy plots of the proposed pNNCLR method vs. the baseline NNCLR, on the Tiny-imagenet and STL-10 datasets. It can be noted that pNNCLR attains comparatively better performance right from the beginning phase of training. The cause for this behavior is also explained in the section \ref{subsec:pnnclr}. While, for both methods, the performance saturates asymptotically, showing a similar trend.

These results indicate that the proposed pNNCLR method performs notably better than the baseline NNCLR method. pNNCLR performs better than other SSL methods on non-medical datasets and comparatively on medical datasets. MoCo performed comparably well with respect to the proposed method.

\begin{table}[t]
    \caption{Effect of modifications in the baseline NNCLR method with smooth-weight-update denoted as (swu), pseudo neighborhood as (pNN), and addition of noise in sampling. Top-1 accuracy is reported for each experiment on the STL-10 dataset.}
    \centering
    \resizebox{.55\linewidth}{!}{
        \begin{tabular}{lc}
            \toprule
            Method                     & Top-1 accuracy ($\uparrow$) \\ \midrule
            Baseline (NNCLR \cite{nnclr})           & 0.7548         \\
            pNNCLR (swu)               & 0.8257         \\
            pNNCLR (swu + pNN)         & 0.8386         \\
            pNNCLR (swu + pNN + noise) & 0.8413         \\ \bottomrule
        \end{tabular}
    }
    \label{tab:ablation variations of pnnclr}
\end{table}

\begin{table}[t]
    \caption{Ablation on the hyperparameters ($\beta$ and $\alpha$) of the proposed ${\rm pNN}(\cdot)$ (pseudo nearest neighbor) sampling approach. Top-1 accuracy is reported on the STL-10 dataset.}
    \centering
    \resizebox{.33\linewidth}{!}{
        \begin{tabular}{lc}
            \toprule
            \multicolumn{2}{c}{Variation in hyperparameters} \\ \midrule
            $\beta$      & Top-1 accuracy      \\ \cmidrule{1-1} \cmidrule(l){2-2}
            0.05                       & 0.8376              \\
            0.10                       & 0.8413              \\ \midrule
            $\alpha$       &       \\ \cmidrule{1-1}
            0.05                       & 0.8286              \\
            0.10                       & 0.8295              \\
            0.15                       & 0.8321              \\
            0.25                       & 0.8386              \\ \bottomrule
        \end{tabular}
    }
    \label{tab:ablation variations in hyperparameters}
\end{table}

\subsection{Ablations}
The ablation study was performed with the baseline NNCLR and proposed method on the STL-10 dataset. In the first ablation, we examined the effect of each proposed modification to the baseline NNCLR method. Table \ref{tab:ablation variations of pnnclr} reports the corresponding results. In that, \textit{swu} denotes smooth weight updation, and \textit{pNN} denotes pseudo nearest neighbor sampling. In Table \ref{tab:ablation variations of pnnclr}, modifying NNCLR with \textit{swu}, pushes the accuracy by $\sim$ 7\%. Modifying NNCLR with (\textit{swu}+\textit{pNN}), pushes the accuracy further by $\sim$ 1\%. Modifying NNCLR with (\textit{swu}+\textit{pNN}+\textit{noise}) further pushes the accuracy to 84.13\%. These results show that \textit{swu} significantly stabilizes the uncertainty in the learning process. Table \ref{tab:ablation variations in hyperparameters} presents variation in the hyperparameters $\alpha$ --- the pseudo sampling control hyperparameter, and $\beta$ --- the noise control hyperparameter. A value of $0.10$ provided a better result over $\alpha = 0.05$. Similarly, $\beta = 0.25$ provided the best results. Table \ref{tab:ablation embedding}, \ref{tab:ablation queue}, \ref{tab:ablation batchsize}, provide results of the ablation study on the embedding size, support-set size and batch size. Table \ref{tab:ablation embedding} shows that the performance of the proposed method increases as the embedding vector size increases. On the other hand, the baseline method's performance maxes out at 2048. For this reason, we used 2048 as the embedding size in all our experiments for all SSL methods. Table \ref{tab:ablation queue} shows that the performance of the proposed method maxes out at a queue size of 10000. Performance of the baseline method maxes out at 20000; however, it is comparable to its performance at 10000. Therefore, we took 10000 as the size of the support set in all our experiments for all SSL methods. Table \ref{tab:ablation batchsize} shows that the proposed and the baseline methods achieve their best performance at a batch size of 64. Therefore, 64 was used as the batch size in all our experiments for all SSL methods.

\begin{table}[t]
    \caption{Effect of using different embedding sizes on the baseline and the proposed method. Top-1 accuracy is reported on the STL-10 dataset.}
    \centering
    \resizebox{.55\linewidth}{!}{
        \begin{tabular}{lcc}
            \toprule
            Embedding size & NNCLR \cite{nnclr}  & Proposed method \\ \midrule
            512            & 0.6061 & 0.7986          \\
            1024           & 0.6580 & 0.8143          \\
            2048           & 0.7548 & 0.8413          \\
            4096           & 0.7537 & 0.8429          \\ \bottomrule
        \end{tabular}
    }
    \label{tab:ablation embedding}
\end{table}

\begin{table}[t]
    \caption{Effect of using different support-set or queue sizes. Top-1 accuracy is reported for the baseline and the proposed method on STL-10 dataset.}
    \centering
    \resizebox{.5\linewidth}{!}{
        \begin{tabular}{lcc}
            \toprule
            Queue size & NNCLR \cite{nnclr}  & Proposed method \\ \midrule
            5000       & 0.7191 & 0.8329          \\
            10000      & 0.7548 & 0.8413          \\
            15000      & 0.7540 & 0.8407          \\
            20000      & 0.7555 & 0.8411          \\ \bottomrule
        \end{tabular}
    }
    \label{tab:ablation queue}
\end{table}

\subsection{Future directions}
In our experiments, we noticed that the size of the support set affects the performance of the SSL method. This has also been observed by other researchers. Conversely, it entails that we need to explore the role of nearest neighbors or pseudo nearest neighbors in other types of SSL methods. We hypothesize that the performance of other SSL methods may benefit from increased diversification of the semantic information. We can also move in the direction of improving the quality of nearest neighbors or the support set as it directly affects the learning in the SSL model. We can also experiment with how effective the features learned by different SSL methods are on the medical image segmentation task. We can also explore the effect of model agnostic variance regularization functions in NNCLR or pNNCLR \cite{vicreg}.

\begin{table}[t!]
    \caption{Effect of using different batch sizes. Top-1 accuracy is reported for the baseline and the proposed method on the STL-10 dataset.}
    \centering
    \resizebox{.5\linewidth}{!}{
        \begin{tabular}{lcc}
            \toprule
            Batch size & NNCLR \cite{nnclr}  & Proposed method \\ \midrule
            16         & 0.7051 & 0.7916          \\
            32         & 0.7141 & 0.8178          \\
            64         & 0.7548 & 0.8413          \\ \bottomrule
        \end{tabular}
    }
    \label{tab:ablation batchsize}
\end{table}

\section{Conclusion}
In this paper, we studied the problem of contrastive self-supervised learning. We covered the development of the field from non-contrastive methods to contrastive methods. It was found that the nearest neighbor sampling based methods were good at increasing semantic variations during unsupervised SSL learning. However, the shortcomings of these methods were also discussed. Further, we proposed pNNCLR, a pseudo nearest neighbor based contrastive learning method, to overcome the weakness of the widely used NNCLR method. We showed how the choice of nearest neighbors in the support set can affect the quality of the learned representations. To avoid this, pNNCLR introduced the use of pseudo nearest neighbors (pNN) with stochastic sampling. Further, a smooth weight updation strategy was also used to stabilize the uncertainty in the learning process. The proposed modifications and multiple recent SSL methods were evaluated on different medical and non-medical standard datasets. Various ablations were performed to fine-tune the hyperparameters. The experiments show that the proposed sampling strategy performs significantly better than the baseline NNCLR approach while competing favorably against the other recent SSL methods.

\appendix

\section{Same class nearest neighbor probability calculation}
\label{app:same class nn probability}
Suppose, the dataset $\mathcal{D}$ on which we are training our SSL network contains classes $c_1, c_2,..., c_{N_c}$, where $N_c$ is the number of classes. Each class $c_i$ has $N_e$ number of items, i.e., we assume that we have a balanced dataset. When choosing a nearest neighbor ${\rm NN}(x_i)$ or ${\rm NN}(z_i)$ for a view of the input $x_i$, from randomly formed support set $Q$ with cardinality $N_q$, the approximate maximal probability, $P[\psi]$ or $P[{\rm class}({\rm NN}(z_i)) = {\rm class}(z_i)]$, can be calculated as follows. Here, $z_i$ is the corresponding embedding of one of the views of $x_i$. We assume that every individual item $x \in \mathcal{D}$ has an equal probability of being randomly selected for forming the support set $Q$. Then, $P[\psi]$ can be defined as:

\begin{equation}
    \label{eq:NN definition}
    \begin{split}
        &P[\psi] = P[{\rm A}\cap{\rm B}] = P[{\rm A}|{\rm B}]\cdot P[{\rm B}] \\
        {\rm where,\;\;} & {\rm A}: {\rm class}({\rm NN}(z_i)) = {\rm class}(z_i),\; {\rm if}\; P[{\rm B}]=1 \\
        &{\rm B}: \big\lvert\{q_i | q_i \in Q \;{\rm and}\; {\rm class}(q_i)={\rm class}(z_i)\}\big\rvert \geq 1,
    \end{split}
\end{equation}

where $\lvert\cdot\rvert$ denotes the set cardinality. In other words, $P[\psi]$ is the probability of the nearest neighbor function ${\rm NN}(\cdot)$ choosing the correct class.
If $\binom{\cdot}{\cdot}$ denotes the binomial coefficient, the probability of the support set $Q$ getting at least one item from the correct class, $P[{\rm B}]$, when items in $Q$ are randomly picked from $\mathcal{D}$ with an equal probability, becomes:

\begin{equation}
    \label{eq:NN probability 1}
    P[{\rm B}] = 1 - P[\,\overline{{\rm B}}\,].
\end{equation}

\begin{equation}
    \label{eq:NN probability 2}
    P[\,\overline{{\rm B}}\,] = \frac{\displaystyle \binom{(N_c-1)N_e}{N_q}}{\displaystyle \binom{N_c N_e}{N_q}}.
\end{equation}

using \eqref{eq:NN probability 2} in \eqref{eq:NN probability 1}, we get

\begin{equation}
    \label{eq:NN probability 3}
    \begin{split}
        P[{\rm B}] &= 1 - \frac{\displaystyle \binom{(N_c-1)N_e}{N_q}}{\displaystyle \binom{N_c N_e}{N_q}} \\
        &= 1 - \frac{(N_c N_e-N_e)!(N_c N_e-N_q)!}{(N_c N_e-N_e-N_q)!(N_c N_e)!} \\
        &= 1 - \frac{N_c N_e-N_q}{N_c N_e}\cdot\frac{N_c N_e-1-N_q}{N_c N_e-1}\cdots\frac{N_c N_e-N_e+1-N_q}{N_c N_e-N_e+1}.
    \end{split}
\end{equation}

\begin{multline}
    \label{eq:NN probability 4}
    \frac{N_c N_e-N_q}{N_c N_e}\cdot\frac{N_c N_e-1-N_q}{N_c N_e-1}\cdots \\ \frac{N_c N_e-N_e+1-N_q}{N_c N_e-N_e+1} < \bigg(\frac{N_c N_e-N_q}{N_c N_e}\bigg)^{N_e}.
\end{multline}

\begin{multline}
    \label{eq:NN probability 5}
    \frac{N_c N_e-N_q}{N_c N_e}\cdot\frac{N_c N_e-1-N_q}{N_c N_e-1}\cdots \\ \frac{N_c N_e-N_e+1-N_q}{N_c N_e-N_e+1} > \bigg(\frac{N_c N_e-N_e+1-N_q}{N_c N_e-N_e+1}\bigg)^{N_e}.
\end{multline}

Using Eq. \eqref{eq:NN probability 3}, Eq. \eqref{eq:NN probability 4} and Eq. \eqref{eq:NN probability 5}, we get a lower and upper bound on $P[{\rm B}]$ as:

\begin{equation}
    \label{eq:NN probability 6}
    1 - \bigg(\frac{N_c N_e-N_q}{N_c N_e}\bigg)^{N_e} < P[{\rm B}] < 1 - \bigg(\frac{N_c N_e-N_e+1-N_q}{N_c N_e-N_e+1}\bigg)^{N_e}.
\end{equation}

Now, let us consider two scenarios, both with a fixed value of $N_q = 10000$. First, when we have a very large dataset where $N_c = 1000$ and $N_e = 1000$, we get $P[{\rm B}] \simeq 0.9999$, using Eq. \eqref{eq:NN probability 6}. Second, when we have a relatively smaller dataset where $N_c = 100$ and $N_e = 100$, or $N_c = 10$ and $N_e = 1000$, we get $P[{\rm B}] \simeq 1$, using Eq. \eqref{eq:NN probability 6}. Hence, the probability $P[{\rm B}]$ stays $\simeq 1$ for both smaller as well as larger size datasets. This changes Eq. \eqref{eq:NN definition} as:

\begin{equation}
    \label{eq:NN probability 7}
    P[\psi] = P[{\rm class}({\rm NN}(z_i)) = {\rm class}(z_i)] = P[{\rm A}],
\end{equation}

which implies that the maximal probability, $P[\psi]$, is proportionate to the probability of choosing the correct class, which in turn depends on the quality of the representations learned by the model $\mathcal{M}_\theta$. As we know that $\mathcal{M}_\theta$ is randomly initialized and stays quite random for the starting epochs, the probability of choosing the correct class becomes $P[\psi] \simeq 0.5$, which is as good as a random selection.

\end{document}